# Convex Analysis of Mixtures for Separating Non-negative Well-grounded Sources

Yitan Zhu, Niya Wang, David J. Miller, *Senior Member, IEEE*, and Yue Wang

*Abstract*— Blind Source Separation (BSS) has proven to be a powerful tool for the analysis of composite patterns in engineering and science. We introduce Convex Analysis of Mixtures (CAM) for separating non-negative well-grounded sources, which learns the mixing matrix by identifying the lateral edges of the convex data scatter plot. We prove a sufficient and necessary condition for identifying the mixing matrix through edge detection, which also serves as the foundation for CAM to be applied not only to the exact-determined and over-determined cases, but also to the under-determined case. We show the optimality of the edge detection strategy, even for cases where source well-groundedness is not strictly satisfied. The CAM algorithm integrates plug-in noise filtering using sector-based clustering, an efficient geometric convex analysis scheme, and stability-based model order selection. We demonstrate the principle of CAM on simulated data and numerically mixed natural images. The superior performance of CAM against a panel of benchmark BSS techniques is demonstrated on numerically mixed gene expression data. We then apply CAM to dissect dynamic contrast-enhanced magnetic resonance imaging data taken from breast tumors and time-course microarray gene expression data derived from in-vivo muscle regeneration in mice, both producing biologically plausible decomposition results.

## I. INTRODUCTION

BLIND Source Separation (BSS) has proven to be a powerful and widely-applicable tool for the analysis of composite patterns in engineering and science, where both source patterns and mixing proportions are of interest but are unknown [1-4]. BSS is often described by a linear latent variable model $\mathbf{X} = \mathbf{AS}$, where $\mathbf{X}$ is the $M \times N$ observation data matrix containing $M$ mixture signals with $N$ data points, $\mathbf{A}$ is the unknown $M \times K$ mixing matrix, and $\mathbf{S}$ is the unknown $K \times N$ source data matrix containing $K$ source signals with $N$ dimensions. The fundamental objective of BSS is to estimate both the unknown mixing proportions and the source signals based only on the observed mixtures.

Over the past fifteen years, a variety of BSS techniques have been continuously reported and tested on synthetic and real data, including Independent Component Analysis (ICA) and its variants, which assume sources are mutually statistically independent or uncorrelated [5-7], and Non-negative Matrix Factorization (NMF) and its variants, which assume mixing proportions and sources are non-negative [1, 8, 9]. NMF is known to have non-unique solutions and can be trapped in a local optimum of its objective function [9]. Efforts, such as the incorporation of sparsity constraints [8, 9], have been made to obtain more well-posed problems under the NMF framework [4, 9-11]. Some other extensions of NMF include relaxation on the signs of the matrix factorization. Semi-NMF allows the mixing matrix to have mixed signs and convex-NMF further requires column vectors in $\mathbf{A}$ to be convex combinations of data points in $\mathbf{X}$ [12]. While these algorithms can usefully extract interesting patterns from mixture observations, they may prove inaccurate or even incorrect in the face of real-world BSS problems, where their pre-imposed assumptions may not be valid. In particular, many source signals are statistically dependent and may not be sparse [2, 3].

Alternative BSS techniques exploit Well-Grounded Points (WGPs) in non-negative source patterns, i.e. points with very high values in one source relative to all other sources [3, 4, 13]. Under the assumption of WGPs, column vectors of the mixing matrix can be estimated by identifying WGPs located at the corners of the mixture observation scatter simplex and, subsequently, the hidden source signals can be recovered. N-FINDR is one of the earliest methods based on WGPs and identifies WGPs by searching for the maximum-volume simplex formed by the data points [14]. Vertex Component Analysis (VCA) implements a fast WGP detection scheme by iteratively projecting data onto a direction orthogonal to the subspace spanned by the WGPs already determined and selecting the data point corresponding to the most extreme projection as the next WGP [15]. The maximum-volume strategy has also been applied in the signal space by nonnegative least-correlated component analysis for recovering well-grounded sources [4]. For cases where WGPs are absent but nearly pure-source data points exist, a

Y. Zhu, N. Wang, and Y. Wang are with the Bradley Department of Electrical and Computer Engineering, Virginia Polytechnic Institute and State University, Arlington, VA 22203, USA (email: yitanzhu@vt.edu; wangny@vt.edu; yuewang@vt.edu).
D. J. Miller is with the Department of Electrical Engineering, Pennsylvania State University, University Park, PA 16802, USA (e-mail: djmiller@engr.psu.edu).
Y. Zhu is also with the Program of Computational Genomics and Medicine, NorthShore University HealthSystem, Evanston, IL 60201, USA (email: yzhu@northshore.org).
Y. Wang is the corresponding author. (e-mail: yuewang@vt.edu; phone: 571-858-3150; address: 900 N. Glebe Road, Arlington, VA 22203, USA)

constrained NMF method considering both the reconstruction error and the minimization of the simplex volume determined by the estimated mixing matrix column vectors has been proposed [16]. A post-processing framework on the results obtained by a WGP-based solution has been developed using either extra mixture data or reliable peak structures of source signals, also for the situations where WGPs are absent [17].

However, there are several potential limitations associated with these techniques. First, they usually lack a theoretical proof of model identifiability and solution optimality [3]. Many methods adopt the strategy of identifying WGPs without a stringent mathematical framework showing its validity. Second, many methods can be applied only to the exact-determined and over-determined cases, where the number of mixtures is no less than the number of sources, but not to the under-determined case, where there are more sources than mixtures. Third, their solutions (including model selection) may be sensitive to noise and outliers in the data. Fourth, some methods do not allow negative elements in the mixing matrix, which limits their applicability.

Based on the realization that the observed pattern across signal indices at each data point can be expressed as a non-negative combination of the column vectors of the mixing matrix [18], we propose a Convex Analysis of Mixtures (CAM) method to estimate the mixing proportions by explicitly identifying WGPs at the lateral edges of the clustered observation scatter plot. CAM is theoretically supported by a series of newly proved identifiability and optimality theorems. A sufficient and necessary condition is discovered for identifying the mixing matrix through edge detection in non-negative well-grounded BSS problems, which also serves as the foundation for CAM to be applied to the under-determined case, in addition to the exact-determined and over-determined cases. The optimality of the edge identification strategy is also proved for non-negative BSS problems, even when WGPs do not exist.

The CAM algorithm integrates a plug-in noise and outlier filtering scheme, an edge detection and geometric convex analysis algorithm, and a model selection scheme for applications on noisy real-world problems. We first design a sector-based clustering scheme, used to obtain an effective noise and outlier-reduced, clustered representation of the data. We then develop an efficient lateral edge detection and geometric convex analysis algorithm that identifies the WGP-associated clusters, whose center vectors are the estimates for the column vectors of the mixing matrix. The algorithm proceeds to estimate source signals by non-negative least-squares fitting of the latent variable model to the observation data, where the number of hidden sources is detected using a stability analysis scheme. Importantly, CAM operates in the scatter space of mixture signals, and hence can address high-dimensional data deconvolution where no observed mixture signal is actually a pure source signal in the signal space. This advantage is significant in that CAM can achieve its goal using only a small number of mixture samples. Moreover, CAM is more powerful in distinguishing between similar sources because it exploits mixing diversity in the scatter space rather than source diversity in the signal space.

We demonstrate the principle and feasibility of the CAM approach on realistic synthetic data involving images and microarray gene expression profiles, and experimentally compare the accuracy of parameter estimates obtained using CAM to the most relevant alternative techniques. We then use the algorithm to dissect dynamic contrast-enhanced magnetic resonance imaging (DCE-MRI) data taken from breast tumors, identifying vascular compartments with distinct pharmacokinetics and revealing intratumor vascular heterogeneity. We also apply CAM to time-course gene expression data derived from in-vivo muscle regeneration in mice, observing biologically plausible dynamic patterns of relevant biological processes with distinct kinetics and phenotype-specific gene expression patterns. Finally, extensions of the CAM algorithm and the relationships to other approaches are discussed.

## II. Theory and Methods

### A. Assumptions of the CAM Model

Considering the linear latent variable model $\mathbf{X} = \mathbf{AS}$, we can re-express the model in vector-matrix notation

$$\mathbf{x}_n = \mathbf{As}_n, \quad \mathbf{A} = [\mathbf{a}_1 ... \mathbf{a}_K], \quad n=1,..., N \quad (1)$$

where $\mathbf{x}_n$, $\mathbf{a}_k$, and $\mathbf{s}_n$ are column vectors of matrices $\mathbf{X}$, $\mathbf{A}$, and $\mathbf{S}$, respectively. Such a linear latent variable model is widely applicable to the analysis of many types of data, with the interpretation of the mixtures and underlying sources application-dependent. As a generic example for now, one can consider image unmixing, with $M$ observed $N$-pixel images, each a mixture of $K$ source images.

Our CAM model is developed based on the following assumptions.

(*A*1) (Source non-negativity) Every element in $\mathbf{S}$ takes a non-negative value and $\mathbf{S}$ has full row rank.

(*A*2) (Sources well-grounded) The source data matrix $\mathbf{S}$ contains at least one WGP on each of the $K$ coordinate axes, i.e. $\forall k \in \{1,...,K\}$, $\exists n_{\text{WGP}(k)}$ such that $\mathbf{s}_{n_{\text{WGP}(k)}} = \lambda \mathbf{e}_k$, $\lambda > 0$, where $\{\mathbf{e}_k\}$ is the standard basis of $K$-dimensional real space.

(*A*3) Every column vector in $\mathbf{A}$ is neither a non-negative nor a non-positive linear combination of other column vectors in $\mathbf{A}$.

(*A*4) $\mathbf{A}$ is of full column rank, i.e. rank$(\mathbf{A})=K$.

(*A*1) is widely satisfied in many real-world applications, such as image mixtures and mixtures of biochemical molecular quantities [2, 3]. (*A*2) exploits the high-contrast segments between source signals, which often exist in many real-world problems [3, 4]. One such example is unmixing of multispectral images in remote sensing, where WGPs correspond to "endmember" pixels. From (*A*1) and Equation (1), we have

$$\mathbf{x}_n = \sum_{k=1}^{K} s_{k,n} \mathbf{a}_k, \quad s_{k,n} \geq 0, \quad n=1,\dots,N \qquad (2)$$

where $s_{k,n}$ is the $k$th element of $\mathbf{s}_n$. When the source matrix $\mathbf{S}$ satisfies (A1) and (A2), i.e. it is a non-negative well-grounded BSS problem, (A3) is a necessary and sufficient condition for the mixing matrix $\mathbf{A}$ to be identified, as we will prove through a set of theorems later. (A4) is a sufficient condition for identifying the source matrix $\mathbf{S}$, when (A1) and (A2) hold, and is widely used in many BSS problems [5]. Apparently, (A3) is a necessary but not a sufficient condition for (A4), in other words, (A3) is guaranteed to hold if (A4) is satisfied, but not vice versa. Also, importantly, (A3) can hold not only in the exact-determined and over-determined cases, but also in the under-determined cases, where there are at least three mixtures, i.e. $M \geq 3$. (A4) on the contrary can be satisfied only in the exact-determined and over-determined cases. (A3) gives CAM the potential to be applied to the under-determined case.

*B. Identifiability of the CAM Model*

We now discuss the identifiability of the CAM model under the aforementioned assumptions via the following definitions and theorems (see formal proofs in Appendices A and B).

**Definition 1.** Given a matrix $\mathbf{B}$ composed by its set of column vectors $\{\mathbf{B}\} = \{\mathbf{b}_1,\dots,\mathbf{b}_Q\}$, the convex cone determined by $\{\mathbf{B}\}$ is

$$\mathcal{C}\{\mathbf{B}\} = \left\{ \sum_{q=1}^{Q} \alpha_q \mathbf{b}_q \,\big|\, \alpha_q \geq 0 \right\} \qquad (3)$$

**Definition 2.** A non-zero vector $\mathbf{z}$ is a lateral edge of $\mathcal{C}\{\mathbf{B}\}$, if $\mathbf{z} \in \mathcal{C}\{\mathbf{B}\}$ (i.e. $\mathbf{z} = \sum_{q=1}^{Q} \alpha_q \mathbf{b}_q,\ \alpha_q \geq 0$) and $\mathbf{z}$ can only be expressed as a trivial combination of $\{\mathbf{B}\}$ (i.e. if $\alpha_q > 0$ for some $q$, then $\mathbf{b}_q = \beta_q \mathbf{z}, \beta_q > 0$).

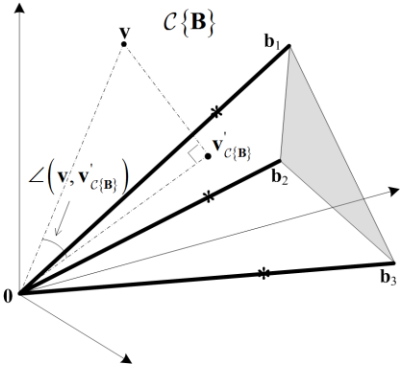

Fig. 1. Illustration of a convex cone $\mathcal{C}\{\mathbf{B}\}$ with three edges in three dimensional space. Lines with an arrow are the axes. Bold lines are edges $\mathbf{b}_1$, $\mathbf{b}_2$ and $\mathbf{b}_3$. The cross-section of convex cone $\mathcal{C}\{\mathbf{B}\}$ is a triangle, indicated by grey color. The star markers on the edges are well-grounded points. $\mathbf{v}$ is a point outside of $\mathcal{C}\{\mathbf{B}\}$. Its projection on $\mathcal{C}\{\mathbf{B}\}$ is $\mathbf{v}'_{\mathcal{C}\{\mathbf{B}\}}$. $\angle(\mathbf{v},\mathbf{v}'_{\mathcal{C}\{\mathbf{B}\}})$ is the projection angle.

See Fig. 1 for illustrations of a convex cone and its lateral edges. Because for edges only the vector direction is of interest, edges with the same vector direction but different lengths will be considered identical in the sequel. With the concept of convex cone, the model assumption (A3) can be formulated as

$$\forall k \in \{1,\dots,K\}, \quad \mathbf{a}_k \notin \mathcal{C}\{\mathbf{A}_{-k}\} \text{ and } \mathbf{a}_k \notin \mathcal{C}\{-\mathbf{A}_{-k}\}$$

where $\mathbf{A}_{-k}$ is the matrix that results from removing the $k$th column from $\mathbf{A}$.

**Lemma 1.** The lateral edges of the convex cone $\mathcal{C}\{\mathbf{A}\} = \left\{ \sum_{k=1}^{K} \alpha_k \mathbf{a}_k \,\big|\, \mathbf{a}_k \in \{\mathbf{A}\}, \alpha_k \geq 0 \right\}$ are the $K$ (mixing matrix) column vectors $\mathbf{a}_1,\dots,\mathbf{a}_K$, if and only if (A3) holds.

**Lemma 2.** Suppose that (A1) and (A2) hold. Then, the convex cone defined by the observed data matrix, i.e. $\mathcal{C}\{\mathbf{X}\} = \left\{ \sum_{n=1}^{N} \alpha_n \mathbf{x}_n \,\big|\, \mathbf{x}_n \in \{\mathbf{X}\}, \alpha_n \geq 0 \right\}$, is identical to $\mathcal{C}\{\mathbf{A}\}$.

**Theorem 1. (Identifiability of the Mixing Matrix).** Suppose that (A1), (A2) hold. The mixing matrix column vectors $\mathbf{a}_1,\dots,\mathbf{a}_K$ can be determined by the lateral edges of $\mathcal{C}\{\mathbf{X}\}$, up to ambiguity of positive scaling and permutation, if and only if (A3) holds.

Theorem 1 is a direct conclusion derived from Lemmas 1 and 2. It states that for separating non-negative well-grounded sources, (A3) is a sufficient and necessary condition for an edge detection solution uniquely identifying the mixing matrix $\mathbf{A}$ based on the observed data $\mathbf{X}$. The *lateral edges* of cone $\mathcal{C}\{\mathbf{X}\}$ are the mixing matrix column vectors $\mathbf{a}_1,\dots,\mathbf{a}_K$. That is, a WGP $\mathbf{x}_{n_{\text{WGP}(k)}} = \sum_{i=1}^{K} s_{i,n_{\text{WGP}(k)}} \mathbf{a}_i = s_{k,n_{\text{WGP}(k)}} \mathbf{a}_k$ is a trivial combination of $\mathbf{a}_1,\dots,\mathbf{a}_K$, it is a lateral edge of cone $\mathcal{C}\{\mathbf{A}\}$, and since $\mathcal{C}\{\mathbf{A}\} = \mathcal{C}\{\mathbf{X}\}$, it is also a lateral edge of cone $\mathcal{C}\{\mathbf{X}\}$. This means that, in principle, under a noise-free scenario, we can directly recover $\mathbf{a}_1,\dots,\mathbf{a}_K$ by locating the lateral edges of $\mathcal{C}\{\mathbf{X}\}$, up to the ambiguity of positive scaling. If (A4) is satisfied, the source data matrix $\mathbf{S}$ can then be recovered by the generalized inverse of $\mathbf{A}$, which is $\mathbf{S} = (\mathbf{A}^T \mathbf{A})^{-1} \mathbf{A}^T \mathbf{X}$ under a noise-free scenario [5]. If (A4) is not satisfied and only (A3) is satisfied, $\mathbf{S}$ might not be recoverable.

We summarize the identifiability of the CAM model as follows:

(1) If (A1), (A2), and (A4) are satisfied, which can happen only in the exact-determined and over-determined cases, both $\mathbf{A}$ and $\mathbf{S}$ are identifiable.

(2) If (A1), (A2), and (A3) are satisfied (which can happen not only in the exact-determined and over-determined cases, but also in the under-determined case where there are at least three mixtures), the mixing matrix $\mathbf{A}$ and the number of sources are identifiable. $\mathbf{S}$ is usually not identifiable if (A4) is not satisfied.

*C. Detectability of the lateral edges of cone $\mathcal{C}\{\mathbf{X}\}$*

One key step in developing the CAM algorithm is efficient detection of the lateral edges of $\mathcal{C}\{\mathbf{X}\}$. Here we discuss the algorithmic principle and optimality of our CAM solution via the following definition and theorems (see formal proofs in Appendices C and D).

**Definition 3.** The projection of a point **v** onto the convex cone $\mathcal{C}\{\mathbf{B}\}$ is

$$\mathbf{v}'_{\mathcal{C}\{\mathbf{B}\}} = \underset{\mathbf{y} \in \mathcal{C}\{\mathbf{B}\}}{\operatorname{argmin}} \|\mathbf{v} - \mathbf{y}\|^2 \qquad (4)$$

Obviously, if $\mathbf{v} \in \mathcal{C}\{\mathbf{B}\}$, then $\mathbf{v}'_{\mathcal{C}\{\mathbf{B}\}} = \mathbf{v}$ and $\angle(\mathbf{v}, \mathbf{v}'_{\mathcal{C}\{\mathbf{B}\}}) = 0$, where $\angle(\cdot, \cdot)$ denotes the angle between two input vectors; if $\mathbf{v} \notin \mathcal{C}\{\mathbf{B}\}$, then $\mathbf{v}'_{\mathcal{C}\{\mathbf{B}\}} \neq \mathbf{v}$ and $\angle(\mathbf{v}, \mathbf{v}'_{\mathcal{C}\{\mathbf{B}\}}) > 0$. We also define the angle between a non-zero vector and a zero vector to equal 180°, i.e. $\angle(\mathbf{v}, \mathbf{0}) = 180°$. See Fig. 1 for an illustration of projecting a data point onto a convex cone and the corresponding projection angle. The optimization problem in Equation (4) is a second order cone programming problem that can be solved by existing algorithms [19].

**Theorem 2 (Property of lateral edges).** Suppose that (*A*1) and (*A*3) hold. Further, assume no two data vectors are in precisely the same direction. Let $\mathbf{x}'_{n, \mathcal{C}\{\mathbf{X}_{-n}\}}$ denote the projection of $\mathbf{x}_n$ onto cone $\mathcal{C}\{\mathbf{X}_{-n}\}$ where $\mathbf{X}_{-n}$ is the data matrix excluding $\mathbf{x}_n$. Then, $\mathbf{x}_n$ is a lateral edge of $\mathcal{C}\{\mathbf{X}\}$, if and only if $\angle(\mathbf{x}_n, \mathbf{x}'_{n, \mathcal{C}\{\mathbf{X}_{-n}\}}) > 0$.

Theorem 2 immediately suggests a simple algorithm for one-by-one detecting all the lateral edges of $\mathcal{C}\{\mathbf{X}\}$, i.e. by applying the angle test of Theorem 2 to check whether $\mathbf{x}_n$ is a lateral edge of cone $\mathcal{C}\{\mathbf{X}\}$, $\forall n$. Note that Theorem 2 assumes each data vector $\mathbf{x}_n$ has a unique direction. This can be easily satisfied in practice by retaining in $\{\mathbf{X}\}$ only one data vector from each group of vectors that are positive scalings of each other (i.e., which lie in the same direction).

An important consideration for the present method is that it requires a WGP to exist for each of the underlying sources. While this is both a reasonable assumption in practice and serves to establish mathematical identifiability of the CAM model, nevertheless in some datasets, WGPs may not exist. It would be helpful to provide an accurate interpretation of the CAM solution in such non-ideal scenarios. Accordingly we show that, even if WGPs do not exist, CAM edge detection provides the *optimal solution* in the sense of capturing maximum source information, because when WGPs are not present, the data vectors which instead achieve *Maximum Source Dominance* (MSD) for each of the sources are the lateral edges of $\mathcal{C}\{\mathbf{X}\}$ and will be identified by CAM. Specifically, we have:

**Theorem 3 (Source dominance optimality).** Suppose that (*A*1) and (*A*4) hold. For each source $k$, $\forall k \in \{1, \ldots, K\}$, the CAM solution identifies at least one lateral edge, denoted by $\mathbf{x}_{n_{\mathrm{MSD}(k)}}$, achieving the maximum source dominance in the sense of

$$\tilde{s}_{k, n_{\mathrm{MSD}(k)}} = \max_{n = 1, \ldots, N} \tilde{s}_{k, n},$$

where $\tilde{\mathbf{s}}_n = [\tilde{s}_{1,n} \ldots \tilde{s}_{k,n} \ldots \tilde{s}_{K,n}]^T$, satisfying $\sum_{k=1}^{K} \tilde{s}_{k,n} = 1$, $\forall n \in \{1, \ldots, N\}$ is the source vector of sample $n$ following a normalization operation applied to the observed data matrix.

Please see Appendix D for the proof of Theorem 3 and for the details of the normalization on the observed data matrix so that source vectors corresponding to different data points are comparable.

### III. CAM ALGORITHM

So far, we have developed a mathematical CAM framework for separating non-negative well-grounded sources under an ideal noise-free situation. In this section, we develop a practical CAM algorithm that is based on this framework but which also robustly addresses the realistic scenarios where there may be both noise and outliers present. This algorithm consists of data preprocessing, sector-based clustering, convex analysis of mixtures, stability analysis, and source pattern recovery. We first summarize the steps of the CAM algorithm and then explain each of these steps in the following subsections.

**CAM Algorithm**

(1) Data preprocessing to normalize data and remove data points with small vector norms that potentially have low local SNR.

(2) Sector-based clustering on the scatter plot to get a noise-reduced representation of the data.

(3) Convex analysis of mixtures for estimating the mixing matrix, including (i) edge detection based on sector central rays to form a candidate pool of estimates for the mixing matrix column vectors and (ii) minimization of model fitting error to produce an estimate for the mixing matrix with a given source number.

(4) Determination of source number by stability analysis, which repeats steps (2) and (3) for different source numbers based on random partitions of the data to calculate the normalized model instability of each candidate source number. The best source number is selected as the one with the smallest instability.

*A. Data Preprocessing*

Our algorithm begins with two data preprocessing steps. First, we scale the observed mixtures to have unit sums and assume the underlying sources also have unit sums as done in [13], i.e. after scaling, $\sum_{n=1}^{N} x_{m,n} = 1$, $\forall m = 1, \ldots, M$, and $\sum_{n=1}^{N} s_{k,n} = 1$, $\forall k = 1, \ldots, K$. Note that this scaling makes each row of **A** have unit sum so that the mixing matrix elements provide the mixing proportions, i.e. $\sum_{k=1}^{K} a_{m,k} = \sum_{k=1}^{K} a_{m,k} \sum_{n=1}^{N} s_{k,n} = \sum_{n=1}^{N} \sum_{k=1}^{K} a_{m,k} s_{k,n} = \sum_{n=1}^{N} x_{m,n} = 1$, $\forall m = 1, \ldots, M$, where $a_{m,k}$ is the $m$th element of $\mathbf{a}_k$. Note that this step removes the scale ambiguity of the BSS solution [6]. Second, consider the following noisy linear latent variable model

$$\mathbf{x}_n = \mathbf{A}\mathbf{s}_n + \boldsymbol{\varepsilon}_n, \quad n = 1, \ldots, N, \qquad (5)$$

where $\boldsymbol{\varepsilon}_n$ is the additive noise on sample $n$ and is independent of $\mathbf{s}_n$. We assume that $\boldsymbol{\varepsilon}_n \sim N(\mathbf{0}, \boldsymbol{\Sigma}_{\text{noise}})$ and define the Signal-to-Noise Ratio (SNR) of the whole dataset as

$$\text{SNR} = \frac{\sum_{n=1}^{N}\|\mathbf{As}_n\|^2}{N \times \text{trace}(\boldsymbol{\Sigma}_{\text{noise}})} \quad (6)$$

Since the expected noise level for all data points is the same, data points with small vector norms are expected to have a lower local SNR, which could have a negative impact on subsequent analysis [3], so the second step of data preprocessing is to remove these small norm points.

### B. Noise or Outlier Removal by Sector-based Clustering

The purpose of sector-based clustering on the preprocessed data points is two-fold: 1) data clustering has proven to be an r model learning [3, 20]; and 2) aggregation of data points into a (smaller) number of clusters improves the computational efficiency of subsequent convex analysis of mixtures by reducing the number of tests performed for identifying lateral edges. After sector-based clustering, each data sector (cluster) is represented by a ray starting from the origin, which is called a sector central ray. Please see Fig. 2 for illustration of sector-based clustering.

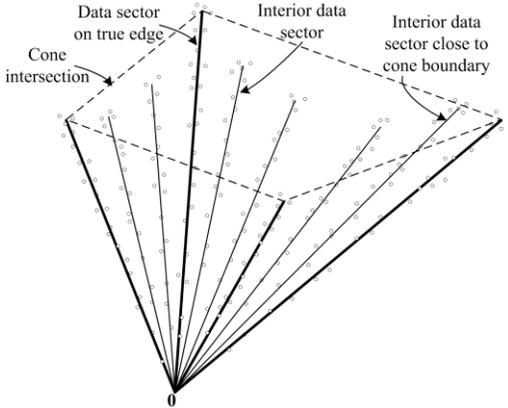

Fig. 2. Illustration of sector-based clustering in a three-dimensional scatter plot. Four sources ($K = 4$) are mixed to form three mixtures ($M = 3$). Small circles are data points. After clustering, each data sector is represented by a sector central ray (solid lines). Four data sectors are close to or on the true edges of cone $\mathcal{C}\{\mathbf{X}\}$, with their sector central rays indicated by bold lines. The quadrilateral formed by the dashed lines indicate the intersection of the cone.

**Definition 4.** The sector central ray $\mathbf{r}_j$ of the $j$th data sector is the ray starting from the origin that minimizes the sum of the squared distances to all the data points in the $j$th data sector.

The distance between a data point and a ray is the minimum distance between the data point and any point on the ray. Sector-based clustering groups data points into sectors (each with its own central ray) so that data points within a sector have more similar orientations (evaluated by their angles made with the central ray) compared to data points in other sectors. Assuming a sufficient number of sectors are used to model the data, we logically impose $\angle(\mathbf{x}_n, \mathbf{r}_j) \leq 90°$, $\forall \mathbf{x}_n \in \psi_j$, $\forall j \in$ $\{1,...,J\}$, where $J$ is the number of data sectors and $\psi_j$ denotes the $j$th data sector. Since only the vector direction is of importance, the sector central rays are confined to have unit norm, i.e. $\|\mathbf{r}_j\| = 1$, $\forall j \in \{1,...,J\}$. Based on Definition 4, the sector central ray is mathematically defined as

$$\mathbf{r}_j = \underset{\|\mathbf{r}\|=1}{\text{argmin}} \sum_{\mathbf{x}_n \in \psi_j} \|\mathbf{x}_n - \mathbf{r}^T\mathbf{x}_n\mathbf{r}\|^2. \quad (7)$$

By expanding the square in the summation and simplifying, we can show that

$$\mathbf{r}_j = \underset{\|\mathbf{r}\|=1}{\text{argmin}} -\mathbf{r}^T\mathbf{C}_j\mathbf{r} = \underset{\|\mathbf{r}\|=1}{\text{argmax}} \mathbf{r}^T\mathbf{C}_j\mathbf{r}. \quad (8)$$

where $\mathbf{C}_j = \sum_{\mathbf{x}_n \in \psi_j} \mathbf{x}_n\mathbf{x}_n^T$ is (the sample-based estimate of) the autocorrelation matrix of data vectors in $\psi_j$. The solution of Equation (8) is the principal eigenvector of $\mathbf{C}_j$.

**Sector-based Data Clustering Algorithm**

(1) Randomly initialize each of the $J$ sector central rays $\mathbf{r}_1,...,\mathbf{r}_J$ to one of the observation data points $\mathbf{x}_1,...,\mathbf{x}_N$ and unit-normalize these vectors.

(2) Partition the observed data points into $J$ data sectors by assigning each data point to its nearest sector based on the distance between the data vector $\mathbf{x}_n$ and the sector central ray $\mathbf{r}$, calculated by $\|\mathbf{x}_n - \mathbf{r}^T\mathbf{x}_n\mathbf{r}\|$.

(3) Update the $J$ sector central rays $\mathbf{r}_1,...,\mathbf{r}_J$ by finding the principal eigenvector of each of the sample-based correlation matrices $\mathbf{C}_j$, $j = 1,...,J$, determined by the data partition in step (2).

(4) Terminate if there is no change in the total clustering distortion shown in Equation (9), from the previous to the current iteration; otherwise, go to step (2).

The sector-based clustering algorithm monotonically descends in the clustering distortion

$$\mathcal{D}(\Psi, \mathbf{R}) = \sum_{j=1}^{J} \sum_{\mathbf{x}_n \in \psi_j} \|\mathbf{x}_n - \mathbf{r}_j^T\mathbf{x}_n\mathbf{r}_j\|^2 \quad (9)$$

where $\mathbf{R} = [\mathbf{r}_1...\mathbf{r}_J]$ is the matrix composed of sector central rays and $\Psi$ is the partition of data points into $J$ data sectors. It also terminates in a finite number of iterations at a fixed point solution that is a local minimum of Equation (9), which can be proved following the standard convergence proof of the generalized Lloyd algorithm [21, 22]. The computational complexity of this algorithm is dominated by the partitioning step, whose complexity is O($JMNI$), where $I$ is the number of algorithm iterations. Random initialization of the sector central rays can affect the local optimum to which the algorithm converges; thus, in practice, the algorithm is usually run multiple times, with the sector partition with the minimum clustering distortion chosen as the final outcome.

## C. Convex Analysis of Mixtures

At this juncture, having performed sector-based clustering, we have $\mathbf{R} = [\mathbf{r}_1 ... \mathbf{r}_J]$ as a noise/outlier mitigated representation of the data matrix $\mathbf{X}$. Accordingly, supported by Theorem 1, which says that in the noise-free case, the columns of $\mathbf{A}$ are the lateral edges of $\mathcal{C}\{\mathbf{X}\}$, it is reasonable, in the noise-mitigated case, to estimate the columns of $\mathbf{A}$ based on the lateral edges of the cone $\mathcal{C}\{\mathbf{R}\}$. CAM uses the following algorithm specifically designed based on Theorem 2 to detect the lateral edges of cone $\mathcal{C}\{\mathbf{R}\}$.

**Cone Lateral Edge Detection Algorithm**

(1) Set $\mathbf{R}_{\text{edge}} = \mathbf{R}$, $j = 1$, and $\tau = 0.001$ (or another small positive value); Set $J^* = J$.

(2) Determine projection image $\mathbf{r}'_{j,\mathcal{C}\{\mathbf{R}_{\text{edge},-j}\}}$ by projecting $\mathbf{r}_j$ onto cone $\mathcal{C}\{\mathbf{R}_{\text{edge},-j}\}$, where $\mathbf{R}_{\text{edge},-j}$ is the matrix resulting from removing the $j$th column from $\mathbf{R}_{\text{edge}}$;

(3) If $\angle(\mathbf{r}_j, \mathbf{r}'_{j,\mathcal{C}\{\mathbf{R}_{\text{edge},-j}\}}) > \tau$, $j = j+1$; otherwise, remove $\mathbf{r}_j$, i.e. the $j$th column, from $\mathbf{R}_{\text{edge}}$ and $J^* = J^* - 1$;

(4) If $j > J^*$, end the algorithm; otherwise, go to step (2).

The worst-case computational complexity of the cone lateral edge detection algorithm is $O(J^3 M)$. After applying the algorithm, the $J^*$ column vectors in $\mathbf{R}_{\text{edge}}$ are the detected edges. The detected edge number has some dependence on the sector-based clustering solution, including the chosen number of sectors, $J$. Clearly, one must choose $J > K$. In practice, to ensure this, one may choose $J$ fairly large, in which case there are usually more than $K$ detected edges. Regardless, the detected edges are good candidates, from which to select a subset as estimates of $\mathbf{a}_1, ..., \mathbf{a}_K$.

To identify good, refined estimates of $\mathbf{a}_1, ..., \mathbf{a}_K$ from this candidate pool, a combinatorial search based on a model fitting error criterion can be performed to identify the most promising $K$ lateral edges. Specifically, let $\{\mathbf{r}_{j_1}, ..., \mathbf{r}_{j_K}\}$ be any size-$K$ subset of $\{\mathbf{R}_{\text{edge}}\}$. The $K$ lateral edges with sector indices $j_1^*, ..., j_K^*$ that minimize a model fitting error are chosen, as follows:

$$(j_1^*, ..., j_K^*) = \underset{(j_1, ..., j_K)}{\operatorname{argmin}} \sum_{j=1}^{J} N_j \angle\left(\mathbf{r}_j, \mathbf{r}'_{j,\mathcal{C}\{\mathbf{r}_{j_1}, ..., \mathbf{r}_{j_K}\}}\right), \quad (10)$$

where $\mathbf{r}'_{j,\mathcal{C}\{\mathbf{r}_{j_1}, ..., \mathbf{r}_{j_K}\}}$ is the projection of $\mathbf{r}_j$ onto cone $\mathcal{C}\{\mathbf{r}_{j_1}, ..., \mathbf{r}_{j_K}\}$ and $N_j$ is the number of data points in sector $j$. Because the angles between the "interior" sector central rays, i.e. sector central rays confined within $\mathcal{C}\{\mathbf{r}_{j_1}, ..., \mathbf{r}_{j_K}\}$, and their projections on $\mathcal{C}\{\mathbf{r}_{j_1}, ..., \mathbf{r}_{j_K}\}$ are all 0, the model fitting error is a weighted sum of the angles between the "exterior" sector central rays and their projections, and the weights are the data sector population sizes. Because this model fitting error is monotonically decreasing as the edge set under consideration enlarges, the search for the best $K$ lateral edges can be accelerated by using the branch and bound search algorithm [23], which guarantees finding the edge set minimizing the model fitting error without the need for exhaustive search. The average complexity of branch and bound search is no larger than $O(\gamma^{J^*-K} JJ^{*2} M)$, where $\gamma > 1$ is a constant that is problem-dependent [41].

The edge set minimizing the model fitting error forms the estimate of the mixing matrix, which we denote by $\widehat{\mathbf{A}}$. We then project all the mixture data vectors in $\mathbf{X}$ onto the cone $\mathcal{C}\{\widehat{\mathbf{A}}\}$ and compose these projected vectors into a matrix, $\mathbf{X}'_{\mathcal{C}\{\widehat{\mathbf{A}}\}}$. This projection step ensures that our estimates for the sources will be non-negative and also helps to suppress noise existing in $\mathbf{X}$. If $\widehat{\mathbf{A}}$ has full column rank, the estimates of sources are calculated via the generalized inverse of $\widehat{\mathbf{A}}$, i.e. $\widehat{\mathbf{S}} = (\widehat{\mathbf{A}}^T \widehat{\mathbf{A}})^{-1} \widehat{\mathbf{A}}^T \mathbf{X}'_{\mathcal{C}\{\widehat{\mathbf{A}}\}}$. Because $\widehat{\mathbf{A}} \widehat{\mathbf{S}} = \mathbf{X}'_{\mathcal{C}\{\widehat{\mathbf{A}}\}}$, which is the projection of $\mathbf{X}$ onto the cone $\mathcal{C}\{\widehat{\mathbf{A}}\}$, it can be shown that $\widehat{\mathbf{S}} = \min_{\overline{\mathbf{S}} \in \mathbb{R}_+^{K \times N}} \|\widehat{\mathbf{A}} \overline{\mathbf{S}} - \mathbf{X}\|_F$, where $\|\cdot\|_F$ denotes the Frobenius norm of a matrix and $\mathbb{R}_+^{K \times N}$ denotes the set of $K$ by $N$ non-negative matrices. Thus, $\widehat{\mathbf{S}}$ is actually a non-negative least squares estimate.

In summary, we note that CAM entails four sub-algorithms that involve minimizing an objective function: 1) sector-based clustering; 2) cone lateral edge detection; 3) estimation of the mixing matrix column vectors through minimization of model fitting error; and 4) stability-based source number estimation. For three of these algorithms, the globally optimal solution that minimizes the given objective function is found. Only the sector-based clustering is subject to finding (potentially poor) locally optimal solutions; however, this is substantially mitigated by performing the clustering multiple times from different random initializations and picking the best solution.

## D. Detection of Source Number by Stability Analysis

One important CAM issue is detection of the structural parameter $K$ (the number of underlying sources), often called model selection. This is indeed particularly critical in real-world applications where the true structure of the latent variable model may be unknown *a priori*. We propose to use a stability analysis scheme to guide model selection, based on a carefully designed model instability index.

Similar to the rationale in determining the number of clusters in data clustering using stability analysis [24], the basic principle is that, if $K$ is too large, some extracted sources will simply model random noise patterns; on the other hand, if $K$ is too small, some extracted sources will be arbitrary combinations of true sources; both scenarios produce unstable models. Stability analysis assesses the model instability indices associated with different values of $K$, calculated based on a large number of 2-fold cross-validations, and selects the model order with lowest model instability. In each cross-validation trial $l \in \{1, ..., L\}$, the preprocessed observation data are randomly divided into two folds (indexed by $l_1$ and $l_2$) of equal size; then, CAM is applied on both folds and produces two independent estimates of the mixing matrix, denoted as $\widehat{\mathbf{A}}_{l_1}(K)$ and $\widehat{\mathbf{A}}_{l_2}(K)$, respectively, for $K = 2,\ldots,K_{\max}$, where $K_{\max}$ is the maximum source number under consideration. We then define the Normalized Model Instability (NMI) index as

$$\text{NMI} = \frac{2\sum_{l=1}^{L} \angle\left(\widehat{\mathbf{A}}_{l_1}(K), \widehat{\mathbf{A}}_{l_2}(K)\right)}{\sum_{l=1}^{L}\left[\angle\left(\widehat{\mathbf{A}}_{l_1}(K), \widehat{\mathbf{A}}_{l_2,\text{rand}}(K)\right) + \angle\left(\widehat{\mathbf{A}}_{l_1,\text{rand}}(K), \widehat{\mathbf{A}}_{l_2}(K)\right)\right]}, \quad (11)$$

where $\widehat{\mathbf{A}}_{l_1,\text{rand}}(K)$ and $\widehat{\mathbf{A}}_{l_2,\text{rand}}(K)$ are estimates of the mixing matrix formed by randomly selecting $K$ sector central rays from the sector-based clustering result obtained on data folds 1 and 2 in the $l$th cross-validation, respectively, and where $\angle(\cdot,\cdot)$ here denotes the minimum average angle between the column vectors of two input matrices. To explicate this averaging, let the two input matrices be $\mathbf{U}=[\mathbf{u}_1\ldots\mathbf{u}_K]$ and $\mathbf{W}=[\mathbf{w}_1\ldots\mathbf{w}_K]$. $\angle(\mathbf{U},\mathbf{W})$ is calculated as

$$\angle(\mathbf{U},\mathbf{W}) = \min_{\boldsymbol{\varphi} \in \boldsymbol{\Phi}_K} \frac{1}{K} \sum_{k=1}^{K} \angle\left(\mathbf{u}_k, \mathbf{w}_{\varphi_k}\right) \quad (12)$$

where $\boldsymbol{\Phi}_K$ is the set including all permutations of $\{1,\ldots,K\}$ and $\varphi_k$ is the $k$th element in a permutation $\boldsymbol{\varphi}$. Since the association between column vectors in $\mathbf{U}$ and $\mathbf{W}$ is not known, we need to search through all possible associations to find the optimal one. Using the Hungarian method, the complexity of this search is $O(K^3)$ [25]. The definition in Equation (11) produces an NMI index that is easy to interpret, and the "normalization" automatically adjusts the NMI index for comparison across different model orders as adopted by [24].

## IV. EXPERIMENTS AND RESULTS

In this section, we first validate the CAM principle on synthetic data and numerically mixed image data, and then show its superior performance against a panel of benchmark BSS techniques on numerically mixed gene expression data under various parameter settings. We evaluate the algorithm performance by comparing the estimates of the mixing matrix and sources to the ground truth, together with the accuracy of source number estimation measured over a number of data set replications. We apply the minimum average angle defined in Equation (12) to assess the accuracy in estimating the true mixing matrix $\mathbf{A}$, via

$$E_{\mathbf{A}} = 1 - \frac{1}{\pi}\angle(\mathbf{A},\widehat{\mathbf{A}}), \quad (13)$$

where $\widehat{\mathbf{A}}$ is the estimate of $\mathbf{A}$. $E_{\mathbf{A}}$ takes a value between 0 and 1, with $E_{\mathbf{A}} = 1$ indicating perfect estimation. The calculation of minimum average angle produces an association between the column vectors in $\widehat{\mathbf{A}}$ and the column vectors in $\mathbf{A}$, which also indicates the association between estimated sources and ground truth sources. To assess the accuracy of source recovery, we use the average correlation coefficient between true sources and their estimates, i.e.

$$E_{\mathbf{S}} = \frac{1}{K}\sum_{k=1}^{K} \rho\left(\mathbf{s}^k, \hat{\mathbf{s}}^k\right), \quad (14)$$

where $\hat{\mathbf{s}}^k$ is the estimate of the $k$th source $\mathbf{s}^k$ that is the $k$th row of source matrix $\mathbf{S}$, and $\rho(\cdot,\cdot)$ denotes the correlation coefficient between them. After validating the CAM principle and showing its superior performance on realistic simulation data, we proceed to apply CAM to real-world scientific problems. Specifically, to dissect DCE-MRI data and real microarray gene expression data, where we evaluate the obtained results against existing biological knowledge.

### A. Demonstration of CAM on Synthetic Data and Numerically Mixed Image Data

To illustrate CAM, we first consider a simulated data set consisting of $N = 1600$ data points. Half of the source vectors are drawn from a three-dimensional exponential distribution with independent variables to ensure the existence of approximate WGPs. The other half are first drawn from a three-dimensional Gaussian distribution with correlated variables to ensure source dependence and then absolute values are taken to force source non-negativity. The mixing matrix, source mean vectors, and covariance matrix are given as

$$\mathbf{A} = \begin{bmatrix} -0.1 & 0.5 & 0.6 \\ 0.6 & -0.1 & 0.5 \\ 0.5 & 0.6 & -0.1 \end{bmatrix}, \boldsymbol{\mu}_{\text{Exp}} = \begin{bmatrix} 1 \\ 1 \\ 1 \end{bmatrix}, \boldsymbol{\mu}_{\text{Gaussian}} = \begin{bmatrix} 0 \\ 0 \\ 0 \end{bmatrix},$$

$$\boldsymbol{\Sigma}_{\text{Gaussian}} = \begin{bmatrix} 1 & 0.9 & 0.9 \\ 0.9 & 1 & 0.9 \\ 0.9 & 0.9 & 1 \end{bmatrix}.$$

The additive noise is drawn from a Gaussian distribution with

$$\boldsymbol{\mu}_{\text{noise}} = \begin{bmatrix} 0 \\ 0 \\ 0 \end{bmatrix}, \quad \boldsymbol{\Sigma}_{\text{noise}} = \begin{bmatrix} 0.07 & 0 & 0 \\ 0 & 0.07 & 0 \\ 0 & 0 & 0.07 \end{bmatrix}.$$

The structure of this data set has been chosen in order to illustrate the noisy and strongly correlated nature of many real data sets. The dataset has an SNR of 12.4dB, calculated by

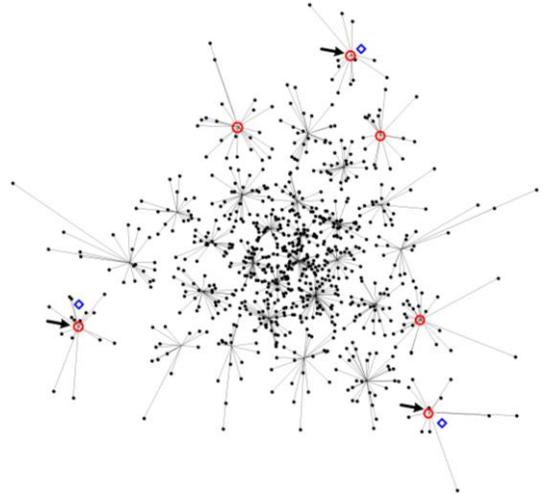

Fig. 3. Perspective projection of the 800 large-norm data points in the toy dataset onto the 2-D intersection of the convex cone formed by the data points. Perspective projection performs simple positive scaling of data points to make every data point have unit element sum. Black dots are data points. Each data point is connected to its sector central ray by a line. Red circles indicate the edges detected by applying the lateral edge detection algorithm on the sector central rays. Blue diamond markers indicate the positions of mixing matrix column vectors. The three edges that minimize the model fitting error among all three-edge sets are indicated by arrows.

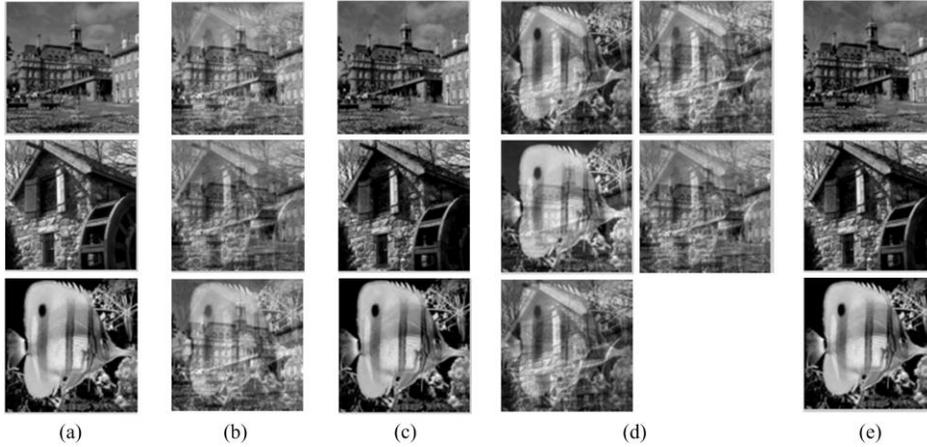

Fig. 4. Separation results of numerically mixed images. (a) Three source images. (b) Three mixture images obtained by mixing the source images in the exact-determined scenario. (c) Recovery result produced by applying CAM to the mixture images in (b). (d) Five mixture images obtained by mixing the source images in (a) in the over-determined scenario. (e) Recovery result produced by applying CAM to the mixture images in (d).

Equation (6).

After data preprocessing, we kept the 800 data points whose vector norms are largest and performed sector-based clustering on these data points 20 times with $J = 30$, selecting the best clustering outcome measured by the total clustering distortion given in Equation (9). On the sector central rays obtained from the best clustering outcome, we performed the cone lateral edge detection algorithm and then identified the three edges that minimized the model fitting error according to Equation (10) to form the estimate of the mixing matrix. The sources were recovered using the mixing matrix estimate accordingly. The resulting recovery accuracies $E_\mathbf{A}$ and $E_\mathbf{S}$ are 0.9826 and 0.9171, respectively. Fig. 3 shows the 800 large-norm data points (black dots), the data sectors and central rays obtained from the best clustering outcome, the detected cone lateral edges (red circles), the three edges producing minimal model fitting error (red circles pointed by arrows), and the ground truth mixing matrix column vectors (blue diamonds). We also applied stability analysis with 30 cross-validations, and obtained NMI indices that show a minimum value at $K = 3$ (see Table 1 for NMI indices of different model orders), which agrees with the ground truth. The power of the CAM approach is supported here as both the mixing matrix and hidden sources are well recovered and the number of hidden sources is correctly identified.

As an example of a realistic problem, we consider data sets containing numerical mixtures of images. Three $103 \times 103$ ($N = 10609$) images (see Fig. 4a) were mixed to produce observation images (see Fig. 4b and Fig. 4d) in both exact-determined ($M = K = 3$) and over-determined ($M = 5 > K = 3$) scenarios, where the randomly generated mixing matrices are

$$\begin{bmatrix} 0.7021 & 0.1506 & 0.1473 \\ 0.5668 & 0.3442 & 0.0890 \\ 0.5535 & 0.1016 & 0.3449 \end{bmatrix} \text{ and } \begin{bmatrix} 0.4082 & 0.3274 & 0.2644 \\ 0.1562 & 0.3085 & 0.5353 \\ 0.3923 & 0.0119 & 0.5958 \\ 0.2376 & 0.4015 & 0.3609 \\ 0.5894 & 0.2941 & 0.1165 \end{bmatrix},$$

respectively. We performed CAM on the mixture image data. Sector-based clustering was run 20 times to select the best clustering outcome, with the sector number set to 70. Stability analysis used 30 cross-validations to detect the number of sources. The source images were quite well recovered and are shown in Fig. 4c and Fig. 4e for the exact-determined and over-determined scenarios, respectively. $E_\mathbf{A}$ and $E_\mathbf{S}$ are 0.9781 and 0.9926, respectively, in the exact-determined scenario. In the over-determined scenario, $E_\mathbf{A}$ and $E_\mathbf{S}$ are 0.9761 and 0.9832, respectively. From Table 1, we can see that the stability analysis based model order selection detects the correct number of source images (i.e. 3) in both the exact-determined and over-determined scenarios.

*B. Performance Comparison on Numerically Mixed Gene Expression Data*

We compared the performance of CAM with six most relevant methods, including non-negative Independent Component Analysis (nICA) [6], Statistical Non-negative Independent Component Analysis (SNICA) [7], Non-negative Matrix Factorization (NMF) [1], Sparse Non-negative Matrix Factorization (SNMF) [8], N-finder algorithm (N-FINDR) [14], and Vertex Component Analysis (VCA) [15].

As a more complex problem, we considered numerical mixtures of four real microarray gene expression profiles ($K = 4$), which are from four distinct ovarian cancer subtypes, i.e. serous, mucinous, endometrioid, and clear cell [26]. The sample labels of the gene expression profiles serving as sources are CHTN-OS-115, UM-OM-001, CHTN-OE-047, and CHTN-OC-033 [26]. The sources contain expression levels of $N = 7069$ genes, some of which are approximately WGPs. The source profiles are highly correlated, with an average pair-wise correlation coefficient of 0.83; also, the source vectors of many genes have very small vector norms. To enable applicability of the NMF methods, we limited mixing matrices to be non-negative. We consider exact-determined ($M = K = 4$), over-determined ($M = 6 > K = 4$), and under-determined ($M = 3 < K = 4$) scenarios, 100 randomly constructed mixing matrices for each scenario, and 6 different SNR levels based on zero-mean white Gaussian additive noise. The mixing matrices are required to have unit row-sums. In the exact-determined and over-determined scenarios, they have a condition number $\leq 4$, so that (A4) holds well. In the under-determined scenario, they satisfy that $\forall \mathbf{a}_k \in \{\mathbf{a}_1,\ldots,\mathbf{a}_K\}$, $\angle(\mathbf{a}_k, \mathbf{a}'_{k,\mathcal{C}\{\mathbf{A}_{-k}\}}) \geq \pi/7$ to ensure that (A3) holds well, where $\mathbf{a}'_{k,\mathcal{C}\{\mathbf{A}_{-k}\}}$ is the projection of $\mathbf{a}_k$ on $\mathcal{C}\{\mathbf{A}_{-k}\}$. To enable the applicability of NMF and SNMF, all observed negative values in data were truncated to 0. In total, there are 1,800 simulation data sets.

TABLE I
NMI INDICES ASSOCIATED WITH DIFFERENT SOURCE NUMBERS OBTAINED WHEN APPLYING CAM ON THE DATASETS

| | Source Number | 2 | 3 | 4 | 5 | 6 | 7 | 8 | 9 |
|---|---|---|---|---|---|---|---|---|---|
| NMI Index | Synthetic data | 0.79 | **0.21** | 0.54 | 0.60 | 0.65 | | | |
| | Image data (exact-determined) | 0.90 | **0.10** | 0.25 | 0.24 | 0.34 | 0.33 | 0.42 | |
| | Image data (over-determined) | 0.45 | **0.08** | 0.38 | 0.23 | 0.41 | 0.48 | 0.44 | 0.50 |
| | DCE-MRI data | 0.39 | **0.29** | 0.58 | 0.64 | 0.65 | 0.71 | 0.70 | 0.78 |
| | Skeletal muscle regeneration gene expression data | 0.51 | 0.71 | **0.45** | 0.69 | 0.73 | 0.74 | 0.79 | 0.82 |

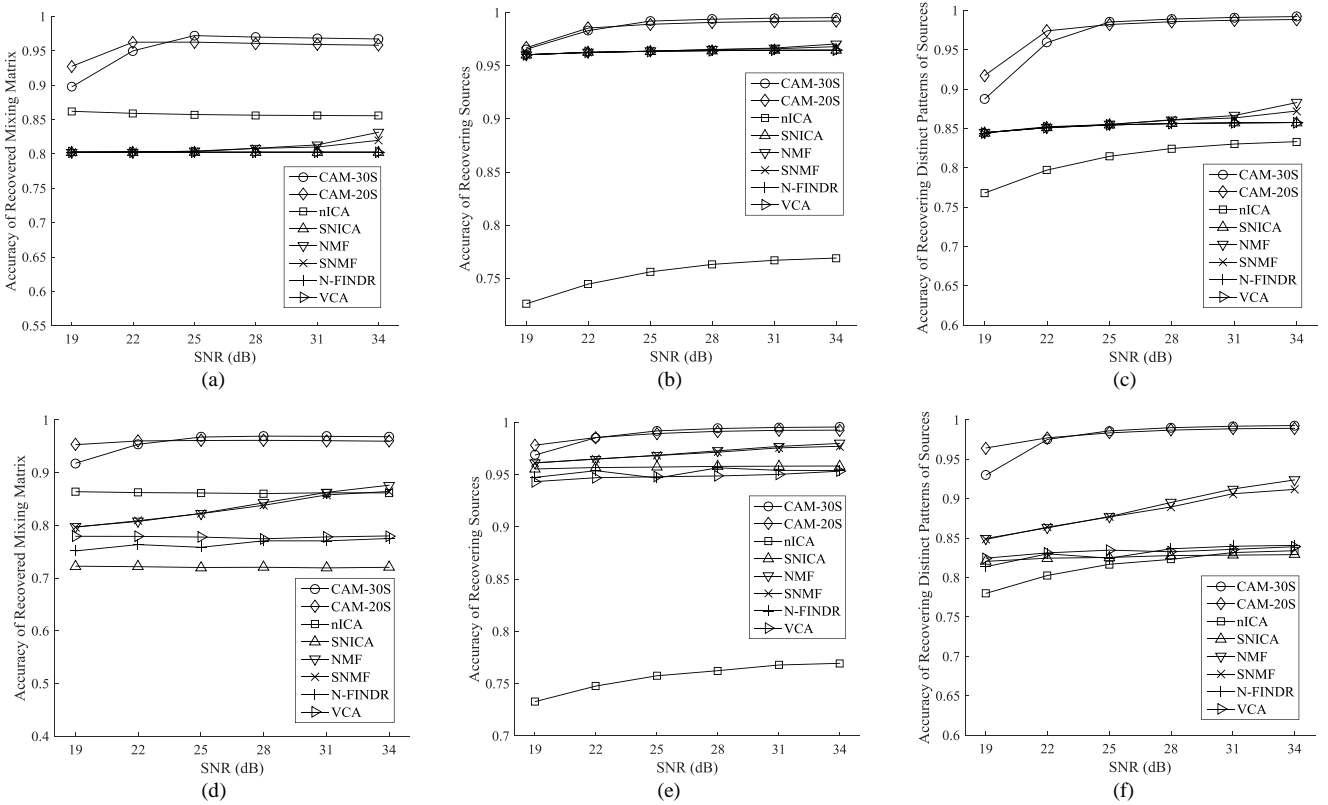

Fig. 5. Performance comparison of CAM and peer methods. (a) and (d) are comparisons on accuracy of recovering the mixing matrix in the exact-determined scenario and over-determined scenario, respectively. (b) and (e) are comparisons on accuracy of recovering sources in the exact-determined scenario and over-determined scenario, respectively. (c) and (f) are comparisons on the accuracy of recovering distinct patterns of sources in the exact-determined scenario and over-determined scenario, respectively.

For CAM, we set the sector numbers $J = 20$ and $J = 30$, with the results indexed by CAM-20S and CAM-30S, respectively. Data preprocessing removed half of the data points with small vector norms. The sector-based clustering always chose the best outcome from 20 independent runs. Stability analysis used 30 cross-validations. We calculated the performance measures for recovering the mixing matrix and the whole gene expression source profiles. More importantly, we also calculated source recovery accuracy over the top source-specific genes -- 800 genes for each ovarian cancer subtype, selected to maximize $s_{k,n}/\sum_{i=1}^{K} s_{i,n}$, $\forall k = 1,\ldots,4$. The distinct source patterns over these genes that are highly expressed in a specific ovarian cancer subtype are of great interest in biological study [27].

When evaluating the accuracies of recovering the mixing matrix, sources, and distinct patterns of sources, the number of sources ($K = 4$) was assumed known and used as an input parameter for all the algorithms. All mixture gene expression profiles were normalized by scaling to have a unit sum before applying CAM and other methods. Principal Component Analysis (PCA) was used to convert an over-determined case to an exact-determined case when applying nICA, SNICA, and N-FINDR in the over-determined experimental scenario [4, 5], because these methods can only work in the exact-determined case. Random initialization was used for setting the initial algorithm parameters needed to run the competing methods. NMF used the multiplicative update rule proposed in [40]. SNMF used the multiplicative update rule proposed in [8], with the source sparseness and model fitting error equally weighted in its objective function. NMF and SNMF terminated when the absolute changes of their objective function values were no larger than 0.0001% or when their numbers of interactions exceeded 5000. The iterative gradient search algorithm of nICA terminated when the mean squared error or its absolute change is smaller than $1\times10^{-9}$ or when the number of interactions exceeded 5000 [6]. SNICA used a simulated annealing algorithm based on constrained Metropolis-type Monte Carlo search to minimize the mutual information between recovered sources [7]. In the initial stage, the Metropolis temperature parameter was set at 0.01, and in the refine stage it was set at $1\times10^{-6}$. The algorithm terminated when the minimum mutual information obtained during the entire run did not decrease in 200 successive Monte Carlo steps. The VCA algorithm requires the SNR to either be estimated or to be input to the algorithm. We found that VCA performance was very poor when the algorithm used its own internal estimation of the SNR. Thus, in our experiments we input the SNR as 100 dB, which basically indicates the data is almost noise-free. This gave more reasonable VCA performance.

Fig. 5 shows the performance results in the exact-determined and over-determined scenarios, when the correct number of sources is given. The estimation accuracies on the mixing matrix, whole hidden sources, and distinct source patterns, are averages over 100 simulation datasets. It can be seen that both CAM-20S and CAM-30S outperform all five peer methods in all cases, and most importantly, they consistently achieve higher accuracy in recovering the distinct source patterns. It should be noted that the use of an overall correlation coefficient in assessing the estimation accuracy of sources may be misleading when the underlying sources are already highly correlated, and the correlation coefficient calculated over the distinct source patterns should be a more meaningful accuracy measure [3]. nICA consistently produced the overall worst unmixing performance among peer methods, which confirms the infeasibility of ICA based approaches to solve BSS problems when the sources are correlated. In all circumstances NMF and SNMF consistently produced similar results. Logically, if the sources are not globally and sufficiently sparse, an insignificant difference between the performances of SNMF and NMF is expected. Though VCA and N-FINDR also exploit the idea of well-grounded sources, they are very sensitive to noise or outliers and thus produce unsatisfactory performance compared to CAM.

To assess the performance of stability based model selection (a unique feature of CAM) at each SNR level, we measured the frequency with which CAM correctly detected the number of sources, over the 100 simulation datasets. Fig. 6 shows this accuracy at different SNR levels in exact-determined and over-determined scenarios. For both CAM-30S and CAM-20S, the number of sources ($K = 4$) was always accurately detected for SNR higher than 25dB in exact-determined and over-determined scenarios. At lower SNR levels, CAM-20S shows a more robust performance against noise than CAM-30S.

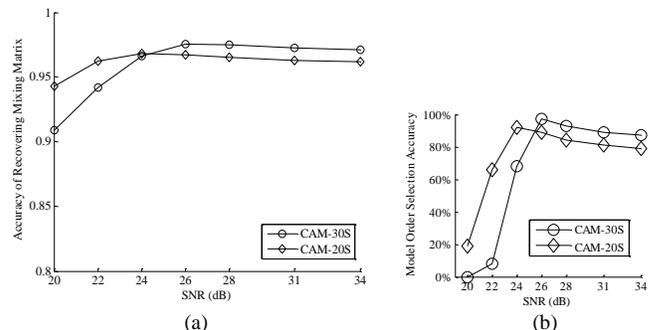

Fig. 7. The performance of CAM on recovering (a) the mixing matrix and (b) the source number in the under-determined scenario.

Fig. 7a shows that CAM can recover the mixing matrix reasonably well over the entire tested SNR range in the under-determined scenario, when the number of sources is given. Fig. 7b shows the accuracy of model order selection in the under-determined scenario, indicating that when the SNR level is higher than 25dB both CAM-20S and CAM-30S detect the correct source number (i.e. 3) on more than 80% of the datasets. In both Fig. 7a and Fig. 7b, some slight performance drop is observed in the under-determined scenario when the SNR is increased toward its high end, possibly due to over-compensation for the noise by the clustering scheme when the noise level is low.

TABLE II
COMPARISON ON EXECUTION TIME (IN SECONDS) OF DIFFERENT METHODS

| Method | Mean | Standard Deviation |
|---|---|---|
| CAM-20S | 24.45 | 2.51 |
| CAM-30S | 33.00 | 4.03 |
| NMF | 7.87 | 4.31 |
| nICA | 1.59 | 0.33 |
| N-FINDR | 5.66 | 0.04 |
| SNMF | 5.15 | 1.97 |
| SNICA | 62.11 | 2.31 |
| VCA | 0.02 | 0.01 |

To compare computational complexity of the methods, we recorded the execution times of all methods, analyzing the 100 datasets for an SNR of 22dB on a computer with a 1.60GHz CPU. The analyses were run with the true number of sources known and with the parameter setting as described above. All methods were implemented in Matlab for a fair comparison, expect for SNICA, which was implemented in C. The mean and standard deviation of execution times in seconds are presented in Table II. VCA is the fastest among all methods, followed by nICA, and then SNMF, N-FINDR and NMF. CAM is slower than these methods, but faster than SNICA, which uses Monte Carlo stochastic search and is the slowest among all competing methods, even with its implementation in C. CAM-30S is slower than CAM-20S as expected, because sector-based clustering takes more time when there are more sectors and the estimation of mixing matrix column vectors through minimization of model fitting error may also take more time due to possibly a larger number of detected edges. We also ran the CAM algorithm with stability analysis based model order selection using 30 cross-validation trials to determine the source number. The mean and standard deviation of execution time for CAM-20S with model order selection are 585.05 seconds and 104.92 seconds, respectively. The mean and standard deviation of execution time for CAM-30S with model order selection are 1019.00 seconds and 161.03 seconds, respectively.

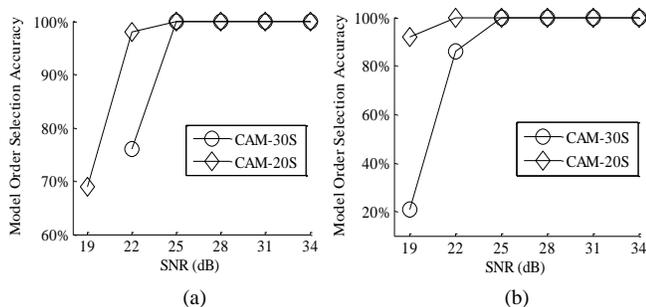

Fig. 6. Model order selection accuracy of CAM. (a) and (b) are the model order selection accuracies obtained in the exact-determined and over-determined scenarios, respectively. The model order selection accuracy of CAM-30S at 19dB is 97% and not drawn in (a), because it is misleading. At 19dB, some of the estimates of mixing matrix obtained by CAM-30S tend to be a permutation and scaling matrix, which indicates poor unmixing. Without effective unmixing, the mixture data dimension is mistaken as the estimated source number that equals the true source number in the exact-determined case, which gives rise to the misleading high model order selection accuracy.

*C. Analysis of Breast Cancer DCE-MRI Data*

As an example of using CAM for real-world application, we considered DCE-MRI data from breast cancer to evaluate tumor vasculature patterns [3]. The data include MRI images of breast tumors taken at sequential time points after the injection of molecular contrast agent into the blood. Due to intratumor heterogeneity and limited imaging resolution, the concentrations of the contrast agent at many image pixels often represent a mixture of more than one vascular compartment, each with distinct and characteristic perfusion and permeability. The existence of near-pure compartment pixels allows us to use CAM to identify distinct vascular compartments and their spatial distributions within a tumor.

The DCE-MRI dataset includes $M = 20$ image frames of a breast tumor (see Fig. 8a) taken every 30 seconds, starting from 90 seconds after injection of the molecular contrast agent. Each image contains $50 \times 50 = 2500$ pixels, and after masking out the non-tumor region, the resulting image contains $N = 715$ pixels for CAM analysis. Noise filtering removed 30% of the pixels whose vector norms were small. The sector-based clustering chose the best clustering outcome in 20 independent runs, with cluster number $J = 30$. We performed stability analysis via 30 cross-validations, which suggested the compartment number $K = 3$, as summarized in Table 1.

CAM analysis indicates three compartments, i.e. fast-flow, slow-flow, and plasma input [28], characterized by their pharmacokinetics patterns. Fig. 8b shows the dynamic changes of tracer concentration of the three compartments, which are the column vectors in the recovered mixing matrix $\hat{\mathbf{A}}$ (with some proper rescaling) [3]. Fig. 8c shows the spatial distributions of the identified compartments, which correspond to the recovered sources $\hat{\mathbf{S}}$.

The fast-flow compartment has a fast tracer clearance rate (see Fig. 8b) and dominates the peripheral "rim" of the tumor (see Fig. 8c). The slow-flow compartment shows very slow tracer kinetics (see Fig. 8b) and dominates the inner "core" of the tumor (see Fig. 8c). The identification of fast-flow and slow-flow pools is plausibly consistent with previously reported intratumor heterogeneity [29, 30]. The defective endothelial barrier function of tumor vessels results in spatially heterogeneous high microvascular permeability to macromolecules [29, 30]. It has been reported that the peripheral "rim" of advanced breast tumors often have active angiogenesis that is essential to tumor development [29]. This rapidly proliferating neovasculature is often abnormal, and forms leaky and chaotic vessels, giving rise to a rapid tracer uptake and washout pattern, forming the fast-flow pool [30]. On the other hand, the inner "core" of the tumor has significantly lower blood flow and oxygen concentration because the tumor growth requires a large portion of its blood supply and also neovessel maturation, forming the slow-flow pool with much slower tracer accumulation and washout [30].

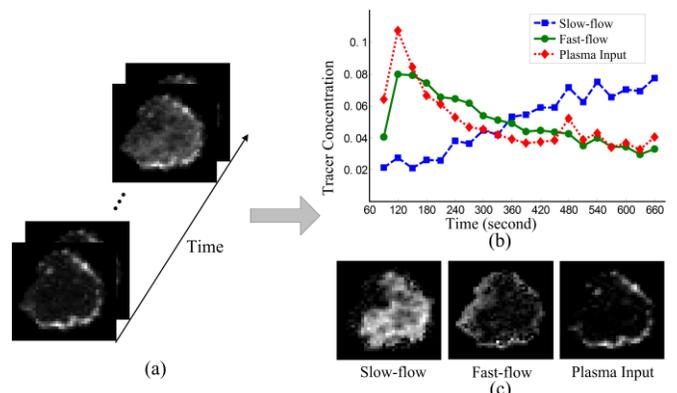

Fig. 8. CAM analysis result on breast cancer DCE-MRI data. (a) MRI images of a breast tumor taken at sequential time points after the injection of molecular contrast agent into blood. (b) Tracer concentration changes of the three identified compartments over time. (c) Recovered source images of the three compartments.

*D. Analysis of Muscle Regeneration Time-Course Gene Expressions*

As a final example, we applied CAM to dissect a time-course gene expression dataset obtained from a mouse skeletal muscle regeneration process [31]. Skeletal muscle regeneration is a highly synchronized process involving the activation of various cellular processes. Cells grow in dynamically evolving subpopulations, yet the dynamics and proportions of cell subpopulations often go unmeasured on the basis of their mRNA expression patterns [32]. Within a mixed population of cells, one might expect distinct cell types to exhibit some distinct patterns of gene expression, and the measured mRNA levels in the mixed cell population represent a weighted average of these hidden biological processes, where the weights are cell proportions involved in different biological processes. Here, we ask whether it is possible to deconvolve the gene expression data from a mixed cell population to discern the proportions of different cell types, by treating specific mRNA patterns as cell-type specific markers [32].

The time-course muscle regeneration gene expression data were acquired at $M = 27$ successive time points using microarrays after the injection of cardiotoxin into the mouse muscle, which damages the muscle tissue and induces staged muscle regeneration [31]. Standard preprocessing suggested $N = 7570$ reliably expressed genes for subsequent CAM analysis [31]. Noise filtering removed 40% of the genes whose vector norms were small. The sector-based clustering chose the best clustering outcome in 20 independent runs, with cluster number $J = 30$. We performed stability analysis via 30 cross-validations, which suggested $K = 4$ as the number of potentially distinct sources associated with underlying active biological processes, as summarized in Table 1.

Fig. 9 displays the source-specific time activity curves (the column vectors of the estimated mixing matrix) that represent the proportions of cell subpopulations associated with the 4 underlying putative biological processes at each time point. For each of the identified sources, we selected 200 source-specific genes (near-WGPs) that maximize $\hat{s}_{k,n}/\sum_{i=1}^{K} \hat{s}_{i,n}$, $\forall k = 1,\ldots,4$, to define source-specific distinct patterns [27]. We input the four source-specific gene groups into Ingenuity Pathway Analysis (IPA), a comprehensive database of gene annotations and functions that performs Fisher's exact test to assess the association of a given gene set with known biological functions, with $p$-values indicating the significance level. Functional analysis by IPA consistently suggests the biological plausibility of all four biological processes revealed by CAM.

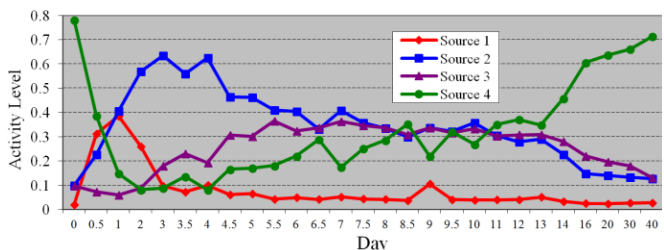

Fig. 9. Time activity curves of the four sources detected on the 27 time-point skeletal muscle regeneration gene expression dataset.

Specifically, IPA suggests that source 1 is associated with inflammation, connective tissue disorders, skeletal and muscular disorders, and immune response, with $p$-values of 6.77E-39, 9.02E-35, 9.02E-35, and 9.62E-32, respectively. The corresponding genes are heavily involved in the necrosis of damaged muscle tissue and the activation of an inflammatory response. In Fig. 9, it can be seen that source 1 activates immediately after muscle damage and then diminishes quickly, reflecting the fact that necrosis and inflammatory response constitute the first transient phase of muscle regeneration [33]. IPA suggests that source 2 is associated with three biological functions, i.e. (1) cell cycle, (2) DNA replication, recombination, and repair, and 3) cellular growth and proliferation, with $p$-values of 7.07E-25, 3.77E-17, and 2.10E-8, respectively. The associated genes are actively involved in myogenic cell proliferation to prepare sufficient myoblasts for later differentiation. The source 2 activity reaches its peak(s) from day 2 to day 4 as biologically expected (see Fig. 9) [33]. IPA suggests that source 3 is associated with tissue development, skeletal and muscular system development, cell to cell signaling and interaction, and connective tissue development and function, with $p$-values of 9.09E-16, 4.91E-11, 2.33E-08, and 4.35E-07, respectively. The corresponding genes are expected to facilitate the differentiation of myoblast into mononucleated myocyte and the fusion of myocytes to form multinucleated myofibers. As expected, in Fig. 9 the source 3 activity goes up after sufficient myoblasts are produced by the activity of source 2, keeps at a high level from day 5 to day 13, and then goes down. Such a trend is consistent with the widely observed fact that muscle regeneration is accomplished in approximately two weeks [33]. IPA suggests that source 4 is associated with skeletal muscular system function and tissue morphology, with a $p$-value of 3.49E-10. The corresponding genes are typically active in normal muscle cells, whose activity drops dramatically after muscle is damaged and gradually recovers until it finally reaches a similar level of original muscular activity as at day 0 (see Fig. 9).

## V. CONCLUSION AND DISCUSSION

We have presented a novel approach to separate non-negative well-grounded sources from observed mixtures, which is geometrically principled and which, as illustrated by real examples, can be very effective at revealing hidden sources within data. It is worth noting that there are four novel features associated with CAM. First, we show both feasibility and optimality of CAM regarding the existence of WGPs and source-dominant points, via newly proved theorems. We prove for the first time a sufficient and necessary condition for identifying the mixing matrix in non-negative well-grounded BSS problems through edge detection. We also show the optimality of the edge detection strategy that identifies the data points with maximum source dominance. Second, we propose an effective noise and outlier removal scheme based on sector-based clustering and an efficient lateral edge detection method on the clustered data scatter plot. Third, based on the proposed identifiability condition of the mixing matrix, the CAM methodology, including the edge detection method and the stability-based model order selection, can be uniformly applied to the exact-determined, over-determined, and under-determined cases, which enables CAM to identify an under-determined problem when encountering such a case. Fourth, we apply CAM to real gene expression data and DCE-MRI data and validate the results against well-established scientific knowledge.

There are important differences between CAM and existing methods for separating non-negative well-grounded sources, such as N-FINDR and VCA [3, 14, 15]. At the front end, these existing methods usually assume the source number is known and apply dimension reduction methods, such as PCA, and the normalization scheme specified in Appendix D on the observed mixture data, so that data points form a simplex in the dimension-reduced space with WGPs being the simplex vertices. The differences between CAM and these simplex-based methods are two-fold. First, CAM does not require prior knowledge of the source number, and uses edge detection and its associated stability-based model order selection to identify the source number. The proposed method can be applied to the exact-determined, over-determined, *and* under-determined problems, as mentioned above, which enlarges the application range of CAM. Second, CAM is *solely* based on convex cone and edge detection, which does not require dimension reduction. Without prior knowledge or accurate estimation of the source number, dimension reduction may *over-reduce* the data dimensionality, with the grave consequence of transforming an exact-determined or over-determined problem into an under-determined one, for which the sources are usually unidentifiable.

A convex model for NMF has been developed in [20], which adopts the source well-groundedness assumption and also uses a clustering method to reduce data points and noise. Compared to this method, CAM has greater applicability by allowing the mixing matrix to have negative elements. Another work related to CAM is [34], which identifies WGPs by examining each observed data point to see whether it is confined within the cone formed by other data points, although the authors do not explicitly formalize their method as an edge detection scheme. Our work enhances the edge detection strategy for separating non-negative well-grounded sources by proving its optimality and identifiability condition. For identifying the mixing matrix in the under-determined case, sparse component analysis proposed in [35] assumes that there are on average $k$ sources contributing to each data point and that $k$ is known *a priori*. [35] uses partial $k$-dimensional subspace clustering to recover the mixing matrix, which can be viewed as an extension of the edge detection strategy from one-source dominant WGPs to $k$-source dominant data points spanning subspaces. CAM allows the mixing matrix to have mixed signs. By requiring both the mixing matrix and sources to be non-negative, methods have been developed to separate well-grounded sources under the NMF framework [36, 37]. The method proposed in [36] identifies WGPs one-by-one by detecting the extreme ray farthest away from the cone formed by the WGPs that have already been detected. [37] proposed a scalable and efficient method for solving problems where $M \gg N$. While CAM is fitting one convex cone to the data, [38] proposed to model the data with multiple small convex cones to accommodate manifold structure in the source signals.

Both probabilistic methods and deterministic methods have been used to solve BSS problems, and there is usually a connection between the two kinds of methods [39]. The proposed CAM method is largely a deterministic approach. It is an interesting topic to build a probabilistic model for separating non-negative well-grounded sources. We are currently investigating a probabilistic CAM model that combines geometric convex analysis with probabilistic modeling. Within a probabilistic modelling framework, information-theoretic criteria, such as minimum description length [3], can be used for model selection to determine the source number. Besides the applications for analyzing genomic data and images, CAM can also be applied to many other analyses, such as document topic modeling [39].

An open-source platform-independent software implementation of the CAM algorithm in R and Java is available from: http://www.cbil.ece.vt.edu/software.htm.

## APPENDIX

### A. Proof of Lemma 1

First, we prove that (*A3*) is a sufficient condition. Suppose that (*A3*) holds. $\forall \mathbf{a}_k \in \{\mathbf{a}_1,...,\mathbf{a}_K\}$, because $\mathbf{a}_k \in \mathcal{C}\{\mathbf{A}\}$, $\mathbf{a}_k$ can be represented by $\mathbf{a}_k = \sum_{j=1}^{K} \alpha_j \mathbf{a}_j$, $\alpha_j \geq 0$, $\forall j \in \{1,...,K\}$. Then we have $(1-\alpha_k)\mathbf{a}_k = \sum_{j=1,j\neq k}^{K} \alpha_j \mathbf{a}_j$, which indicates $\alpha_k = 1$, because otherwise $\mathbf{a}_k \in \mathcal{C}\{\mathbf{A}_{-k}\}$ or $\mathbf{a}_k \in \mathcal{C}\{-\mathbf{A}_{-k}\}$, and (*A3*) is violated. Thus, $\sum_{j=1,j\neq k}^{K} \alpha_j \mathbf{a}_j = \mathbf{0}$. Because (*A3*) holds, $\forall j \neq k$, $\mathbf{a}_j \notin \mathcal{C}\{-\mathbf{A}_{-(k,j)}\} \subseteq \mathcal{C}\{-\mathbf{A}_{-j}\} \Rightarrow \alpha_j = 0$, where $\mathbf{A}_{-(k,j)}$ is the matrix resulting from removing the $k$th and $j$th columns from $\mathbf{A}$. This indicates that $\mathbf{a}_k$ must be a trivial non-negative combination of $\{\mathbf{A}\}$ and thus $\mathbf{a}_k$ is an edge.

Second, we prove that (*A3*) is a necessary condition. Suppose that (*A3*) is not satisfied. Then, $\exists k \in \{1,...,K\}$, $\mathbf{a}_k \in \mathcal{C}\{\mathbf{A}_{-k}\}$ or $\mathbf{a}_k \in \mathcal{C}\{-\mathbf{A}_{-k}\}$. Also, $\mathbf{a}_k \neq \gamma \mathbf{a}_j$, $\forall j \in \{1,...,K\}$, $j \neq k$, and $\forall \gamma > 0$, because otherwise the model is degenerate. $\mathbf{a}_k \in \mathcal{C}\{\mathbf{A}_{-k}\} \Rightarrow \mathbf{a}_k = \sum_{j=1,j\neq k}^{K} \alpha_j \mathbf{a}_j$, $\alpha_j \geq 0$, $\forall j \in \{1,...,K\}$ and $j \neq k$. Because $\mathbf{a}_k \neq \mathbf{0}$, $\exists \alpha_j > 0$ and $j \neq k$. We can represent $\mathbf{a}_k$ by $\mathbf{a}_k = \mathbf{a}_k/2 + \sum_{j=1,j\neq k}^{K}(\alpha_j/2)\mathbf{a}_j$. Thus $\mathbf{a}_k$ is a non-trivial combination of $\{\mathbf{A}\}$ and is not an edge. In a similar way, we can show that $\mathbf{a}_k \in \mathcal{C}\{-\mathbf{A}_{-k}\}$ also makes $\mathbf{a}_k$ a non-trivial combination of $\{\mathbf{A}\}$ and thus is not an edge. This indicates that (*A3*) must be satisfied for $\{\mathbf{a}_1,...,\mathbf{a}_K\}$ to be the edges.

### B. Proof of Lemma 2

Any vector $\mathbf{v} \in \mathcal{C}\{\mathbf{X}\}$ can be represented by $\mathbf{v} = \sum_{n=1}^{N} \alpha_n \mathbf{x}_n = \sum_{n=1}^{N} \alpha_n \mathbf{A} \mathbf{s}_n = \mathbf{A} \sum_{n=1}^{N} \alpha_n \mathbf{s}_n$, where $\alpha_n \geq 0$, $\forall n \in \{1,...,N\}$. Moreover, $\sum_{n=1}^{N} \alpha_n \mathbf{s}_n$ is a $K$ dimensional non-negative vector. Therefore, $\mathbf{v} \in \mathcal{C}\{\mathbf{A}\}$, and we have proved $\mathcal{C}\{\mathbf{X}\} \subseteq \mathcal{C}\{\mathbf{A}\}$.

Let $\{n_{\text{WGP}(1)},...,n_{\text{WGP}(K)}\}$ be the indices of a WGP set, where $\mathbf{x}_{n_{\text{WGP}(k)}}$ is a WGP of source $k$. Any vector $\mathbf{v} \in \mathcal{C}\{\mathbf{A}\}$ can be represented by

$$\mathbf{v} = \sum_{k=1}^{K} \alpha_k \mathbf{a}_k = \sum_{k=1}^{K} \frac{\alpha_k}{s_{k,n_{\text{WGP}(k)}}} \mathbf{x}_{n_{\text{WGP}(k)}},$$

where $\alpha_k \geq 0$, $\forall k \in \{1,...,K\}$. Obviously, $\alpha_k/s_{k,n_{\text{WGP}(k)}} \geq 0$, so $\mathbf{v} \in \mathcal{C}\{\mathbf{X}\}$, and we have proved $\mathcal{C}\{\mathbf{A}\} \subseteq \mathcal{C}\{\mathbf{X}\}$.

Therefore, $\mathcal{C}\{\mathbf{A}\} = \mathcal{C}\{\mathbf{X}\}$.

### C. Proof of Theorem 2

First, we assume that $\angle(\mathbf{x}_n, \mathbf{x}'_{n,\mathcal{C}\{\mathbf{X}_{-n}\}}) > 0$, which means $\mathbf{x}_n \notin \mathcal{C}\{\mathbf{X}_{-n}\}$. Because $\mathbf{x}_n \in \mathcal{C}\{\mathbf{X}\}$, we can write $\mathbf{x}_n = \sum_{i=1}^{N} \alpha_i \mathbf{x}_i \Leftrightarrow (1-\alpha_n)\mathbf{x}_n = \sum_{i=1,i\neq n}^{N} \alpha_i \mathbf{x}_i$, where $\alpha_i \geq 0$, $\forall i \in \{1,...,N\}$. Because $\mathbf{x}_n \notin \mathcal{C}\{\mathbf{X}_{-n}\}$, $\alpha_n \geq 1$. We can further write $\mathbf{0} = \sum_{i=1,i\neq n}^{N} \alpha_i \mathbf{x}_i + (\alpha_n - 1)\mathbf{x}_n = \mathbf{A}(\sum_{i=1,i\neq n}^{N} \alpha_i \mathbf{s}_i + (\alpha_n - 1)\mathbf{s}_n)$, where $\sum_{i=1,i\neq n}^{N} \alpha_i \mathbf{s}_i + (\alpha_n - 1)\mathbf{s}_n$ is a non-negative vector. Actually, it must be a zero vector, because otherwise (*A3*) is violated. Because (*A1*) is satisfied, $\mathbf{s}_i$ is a non-negative, non-zero vector, $\forall i \in \{1,...,N\}$. Then we must have $\alpha_i = 0$, $\forall i \neq n$, and $\alpha_n = 1$. So $\mathbf{x}_n$ can only be a trivial non-negative combination of $\mathbf{x}_1,...,\mathbf{x}_N$, which means that $\mathbf{x}_n$ is a lateral edge of $\mathcal{C}\{\mathbf{X}\}$.

Second, suppose that $\angle\left(\mathbf{x}_n, \mathbf{x}'_{n,\mathcal{C}\{\mathbf{X}_{-n}\}}\right) = 0$, which means $\mathbf{x}_n \in \mathcal{C}\{\mathbf{X}_{-n}\}$. Also, for simplicity of discussion, assume that $\mathbf{x}_1,...,\mathbf{x}_N$ have different vector directions, i.e. no vector is a positive scaling of another vector. $\mathbf{x}_n$ can be represented by $\mathbf{x}_n = \sum_{i=1, i \neq n}^{N} \alpha_i \mathbf{x}_i$, where $\alpha_i \geq 0$, $\forall i \neq n$, and $\alpha_i > 0$ for at least two data points other than $\mathbf{x}_n$. Thus we can write $\mathbf{x}_n$ as a non-trivial non-negative combination of $\mathbf{x}_1,...,\mathbf{x}_N$, for example, $\mathbf{x}_n = \mathbf{x}_n/2 + \sum_{i=1, i\neq n}^N (\alpha_i/2)\mathbf{x}_i$, which means that $\mathbf{x}_n$ is not a lateral edge of $\mathcal{C}\{\mathbf{X}\}$. Therefore, $\mathbf{x}_n$ can be a lateral edge of $\mathcal{C}\{\mathbf{X}\}$ only if $\angle\left(\mathbf{x}_n, \mathbf{x}'_{n,\mathcal{C}\{\mathbf{X}_{-n}\}}\right) > 0$.

### D. Proof of Theorem 3

We define maximum source dominance with respect to a normalized version of the data points. Because $\mathbf{A}$ has full column rank, there exist $K$ linearly independent rows in $\mathbf{A}$ forming a basis, i.e. any real-valued $K$-dimensional vector can be formed by linear combination of the $K$ row vectors in $\mathbf{A}$, including any non-negative vectors. Thus, there must exist a vector $\boldsymbol{\gamma}$ satisfying $\boldsymbol{\gamma}^T \mathbf{a}_k > 0$, $\forall k = 1,...,K$. The mixture data vectors are scaled to have unit inner products with $\boldsymbol{\gamma}$, i.e.

$$\tilde{\mathbf{x}}_n = \frac{\mathbf{x}_n}{\boldsymbol{\gamma}^T \mathbf{x}_n} = \frac{\mathbf{A}\mathbf{s}_n}{\boldsymbol{\gamma}^T \mathbf{x}_n} = \left[\frac{\mathbf{a}_1}{\boldsymbol{\gamma}^T \mathbf{a}_1}, ..., \frac{\mathbf{a}_K}{\boldsymbol{\gamma}^T \mathbf{a}_K}\right] \begin{bmatrix} \frac{\boldsymbol{\gamma}^T \mathbf{a}_1 s_{1,n}}{\boldsymbol{\gamma}^T \mathbf{x}_n} \\ \vdots \\ \frac{\boldsymbol{\gamma}^T \mathbf{a}_K s_{K,n}}{\boldsymbol{\gamma}^T \mathbf{x}_n} \end{bmatrix} = \tilde{\mathbf{A}}\tilde{\mathbf{s}}_n,$$

where $\tilde{\mathbf{s}}_n = [\boldsymbol{\gamma}^T\mathbf{a}_1 s_{1,n}/\boldsymbol{\gamma}^T\mathbf{x}_n \ldots \boldsymbol{\gamma}^T\mathbf{a}_K s_{K,n}/\boldsymbol{\gamma}^T\mathbf{x}_n]^T$ and $\tilde{\mathbf{A}} = [\mathbf{a}_1/\boldsymbol{\gamma}^T\mathbf{a}_1, ..., \mathbf{a}_K/\boldsymbol{\gamma}^T\mathbf{a}_K]$. Obviously, $\tilde{s}_{k,n} = \boldsymbol{\gamma}^T\mathbf{a}_k s_{k,n}/\boldsymbol{\gamma}^T\mathbf{x}_n \geq 0$, $\forall k \in \{1,...,K\}$, and $\sum_{k=1}^K \tilde{s}_{k,n} = 1$. $\tilde{s}_{k,n}$ defines the level/abundance of source $k$ in the $n$th data point after normalization. Because the normalization only performs a positive scaling of the data vectors, the lateral edges of $\mathcal{C}\{\tilde{\mathbf{X}}\}$ remain the same as those of $\mathcal{C}\{\mathbf{X}\}$ and thus can be identified by the edge detection strategy implied by Theorem 2.

Consider $\tilde{\mathbf{x}}_{n_k^*}$, whose $k$th source abundance is the largest, i.e. such that $n_k^* = \arg\max_{n=1,...,N} \tilde{s}_{k,n}$,

$$\tilde{\mathbf{x}}_{n_k^*} = \frac{\mathbf{x}_{n_k^*}}{\boldsymbol{\gamma}^T\mathbf{x}_{n_k^*}} = \frac{\sum_{n=1}^N \alpha_n \mathbf{x}_n}{\boldsymbol{\gamma}^T \sum_{n=1}^N \alpha_n \mathbf{x}_n}$$
$$= \sum_{n=1}^N \frac{\alpha_n \boldsymbol{\gamma}^T \mathbf{x}_n}{\sum_{i=1}^N \alpha_i \boldsymbol{\gamma}^T \mathbf{x}_i} \frac{\mathbf{x}_n}{\boldsymbol{\gamma}^T \mathbf{x}_n} = \sum_{n=1}^N \tilde{\alpha}_n \tilde{\mathbf{x}}_n,$$

where $\alpha_n \geq 0$, $\forall n \in \{1,...,N\}$, and $\tilde{\alpha}_n = \alpha_n \boldsymbol{\gamma}^T \mathbf{x}_n / \sum_{i=1}^N \alpha_i \boldsymbol{\gamma}^T \mathbf{x}_i$. Obviously, $\tilde{\alpha}_n \geq 0$ and $\sum_{n=1}^N \tilde{\alpha}_n = 1$. Because $\tilde{\mathbf{x}}_{n_k^*} = \sum_{n=1}^N \tilde{\alpha}_n \tilde{\mathbf{x}}_n$, we can write

$$\tilde{\mathbf{x}}_{n_k^*} = \sum_{j=1}^K \tilde{\mathbf{a}}_j \tilde{s}_{j,n_k^*} = \sum_{n=1}^N \tilde{\alpha}_n \sum_{j=1}^K \tilde{\mathbf{a}}_j \tilde{s}_{j,n}$$
$$\Leftrightarrow \sum_{j=1}^K \tilde{\mathbf{a}}_j \left(\tilde{s}_{j,n_k^*} - \sum_{n=1}^N \tilde{\alpha}_n \tilde{s}_{j,n}\right) = 0.$$

Because $\tilde{\mathbf{a}}_1,...,\tilde{\mathbf{a}}_K$ are linearly independent, $\tilde{s}_{k,n_k^*} - \sum_{n=1}^N \tilde{\alpha}_n \tilde{s}_{k,n} = \sum_{n=1}^N \tilde{\alpha}_n \left(\tilde{s}_{k,n_k^*} - \tilde{s}_{k,n}\right) = 0$. Define $\Delta_< = \{n | \tilde{s}_{k,n} < \tilde{s}_{k,n_k^*}\}$ and $\Delta_= = \{n | \tilde{s}_{k,n} = \tilde{s}_{k,n_k^*}\}$. We have $\tilde{\alpha}_n = 0$, $\forall n \in \Delta_<$, and thus $\sum_{n \in \Delta_=} \tilde{\alpha}_n = 1$. So, $\tilde{\mathbf{x}}_{n_k^*}$ lies within a convex hull formed by $\{\tilde{\mathbf{x}}_n | n \in \Delta_=\}$. Consider a vertex of this convex hull, denoted by $\tilde{\mathbf{x}}_{n_k^*}^{**}$, which also achieves the maximum dominance of source $k$. Because $\tilde{\mathbf{x}}_{n_k^*}^{**} \in \mathcal{C}\{\tilde{\mathbf{X}}\}$ and based on the above derivation, we must have $\tilde{\mathbf{x}}_{n_k^*}^{**} = \sum_{n=1}^N \tilde{\beta}_n \tilde{\mathbf{x}}_n$, $\tilde{\beta}_n \geq 0$, $\sum_{n \in \Delta_=} \tilde{\beta}_n = 1$, and $\tilde{\beta}_n = 0$, $\forall n \in \Delta_<$. Because $\tilde{\mathbf{x}}_{n_k^*}^{**}$ is a vertex of the convex hull, it can only be a trivial combination of $\{\tilde{\mathbf{x}}_n | n \in \Delta_=\}$ (i.e. if $\tilde{\beta}_n > 0$ for any $n \in \Delta_=$, then $\tilde{\mathbf{x}}_{n_k^*}^{**} = \tilde{\mathbf{x}}_n$), which indicates that $\tilde{\mathbf{x}}_{n_k^*}^{**}$ can only be a trivial combination of $\{\tilde{\mathbf{X}}\}$. Thus $\tilde{\mathbf{x}}_{n_k^*}^{**}$ is a lateral edge of $\mathcal{C}\{\tilde{\mathbf{X}}\}$ and $\mathcal{C}\{\mathbf{X}\}$, and can be identified by the CAM solution. Note that $\Delta_= = \{n_k^*\}$ is a special case, wherein the convex hull reduces to a single point vertex, and in such a case $\tilde{\mathbf{x}}_{n_k^*}$ is a lateral edge identified by the CAM solution.

ACKNOWLEDGMENT

The authors would like to thank Eric P. Hoffman of the Children's National Medical Center and Peter L. Choyke of the National Cancer Institute for providing biomedical data and expert advice.